\documentclass[letterpaper, 10 pt, conference]{ieeeconf}

\IEEEoverridecommandlockouts
\usepackage{cite}
\usepackage{amsmath,amssymb,amsfonts}
\usepackage{cite}
\usepackage{graphicx}
\usepackage{textcomp}
\usepackage{balance}
\usepackage[hyphens]{url}  \usepackage{hyperref}
\usepackage{booktabs}
\usepackage[table,xcdraw]{xcolor}
\usepackage{multirow}
\usepackage{multicol}
\usepackage{comment}
\usepackage{algorithm}
\usepackage{algpseudocode}

\algrenewcommand{\EndIf}{\textcolor{white}{.}}
\algrenewcommand{\EndWhile}{\textcolor{white}{.}}

\begin{document}

\title{\LARGE \bf
Social and Telepresence Robots for  Accessibility and Inclusion in  Small Museums}

\author{Nello Balossino$^{1}$ $^{4}$, Rossana Damiano$^{1}$, Cristina Gena$^{1}$, Alberto Lillo$^{1}$, Anna Maria Marras$^{3}$, \\ Claudio Mattutino$^{1}$, Antonio Pizzo$^{2}$, Alessia Prin$^{5}$, Fabiana Vernero$^{1}$
\thanks{$^{1}$Department of Computer Science, University of Turin,
        Italy {\tt\small name.surname@unito.it}}%
\thanks{$^{2}$Department of Humanities, University of Turin,
        Italy {\tt\small name.surname@unito.it}}%
        \thanks{$^{3}$Department of Historical Studies, University of Turin,
        Italy {\tt\small name.surname@unito.it}}%
        \thanks{$^{4}$Museum of the Holy Shroud, Turin,
        Italy
        }%
        \thanks{$^{5}$Unione Montana Comuni Olimpici Via Lattea, Cesana Torinese,
        Italy
        }%
}

\maketitle
\thispagestyle{empty}
\pagestyle{empty}

\begin{abstract}
There are still many  museums that present accessibility barriers, particularly regarding perceptual, cultural, and cognitive aspects. This is especially evident in low-density population areas. The aim of the ROBSO-PM project  is to to improve the accessibility of small museums  with the use of social robots and social telepresence robots, focusing on three museums as a case study:  the Museum of the Holy Shroud in Turin, a small but globally known institution,  and two lesser-known mountain museums, the Museum of the Champlas du Col Carnival, and the Pragelato Museum of Alpine Peoples' Costumes and Traditions. The project explores two main applications for robots: as guides to support inclusive visits for foreign or disabled visitors, and as telepresence tools allowing people with limited mobility to access museums remotely. From a research perspective, key topics include storytelling, robot personality, empathy, personalization, and, in the case of telepresence, collaboration  between the robot and the person, with clearly defined roles and autonomy.
\end{abstract}

\section{Introduction and context}

Museums represent controlled but dynamic environments, ideal for testing advanced robotic technologies. In this context, both autonomous and telepresence robots have been employed in the past to enhance visitor experience and stimulate new forms of social and cultural interaction.

Over the past thirty years, museum robotics has made significant progress.
From early experiments like Polly \cite{horswill1993polly}, developments have led to increasingly sophisticated systems such as Rhino \cite{burgard1999rhino}, Minerva \cite{thrun2000minerva}, and Chips, up to RoboX \cite{arras2003robox}, which introduced an emotional state machine. More recently, R1, developed by IIT and tested at the GAM in Turin in 2022 \cite{iit2022r1}, represents an advanced synthesis of artificial intelligence, multimodal communication, and adaptability.

Robots in museums are designed to perform several functions: guiding visitors, providing information about exhibits, assisting with navigation through exhibition spaces, offering general support, and maintaining engagement through social behaviors such as gestures, facial expressions, and natural language \cite{jensen2005robox}. However, to be effective, such robots must meet specific technical requirements: ensuring safety through collision avoidance systems and emergency stop buttons \cite{burgard1999rhino}; autonomously planning and adapting their movement in complex environments; and functioning as cognitive agents \cite{macaluso2005cicerobot}. They must also detect faults and request assistance when needed. The quality of interaction is crucial—studies show that users value interactive capabilities even more than navigation \cite{thrun2000minerva}. Challenges persist, including ambient noise, uncooperative users, and short interaction times \cite{jensen2005robox, kanda2003practical}.

Telepresence robotics has added new perspectives, enabling immersive remote participation. Initially used in business settings, it is now applied in healthcare, education, and museums \cite{casoni2020telepresence}. Devices like Virgil \cite{giuliano2017virgil} and Rhino's remote interface illustrate how such robots can serve as cultural mediators. These systems allow users to control the robot remotely and interact with on-site visitors, enhancing accessibility and social presence. However, embodiment remains limited and asymmetrical \cite{cha2017embodiment}, and current research is exploring how to make the experience more emotionally expressive and interactive.

Within this scenario, the ROBSO-PM project introduces the Sanbot robot in the Museum of the Holy Shroud in Turin, a small but globally known institution,  and in two lesser-known mountain museums, the Museum of the Champlas du Col Carnival, and the Pragelato Museum of Alpine Peoples' Costumes and Traditions. We aim to activate a co-design process with the museum to  make the visit accessible to all visitors regardless of physical or cognitive ability and regardless of their physical presence. Sanbot will operate both as a guide and as a telepresence interface, enabling inclusive field experimentation. In pursuing this goal, we do not intend for the robot to be used in the same way as conventional telepresence robots, which typically serve as embodied extensions of the remote user controlling them. Our goal is for Sanbot to retain its own identity and to actively engage with visitors by guiding them through the museum, offering its own perspective, physical presence, and kinetic expressiveness. To achieve this, we will introduce a new form of interaction between the human and the robot that must be co-designed and evaluated together with museum visitors.

This initiative supports collaboration between academia, cultural institutions, and local stakeholders. Our goal is to explore the potential of robotics in promoting small museums and fostering cultural development. Future directions include IoT integration \cite{kildal2018iot}, stereotypical user and group modeling \cite{GenaArdissono2001, masthoff2004group}, empathic interaction \cite{trejo2018group}, and multimodal design strategies \cite{schmidt2020beyond}. By adopting a human-centered perspective, robotics in museums can contribute to creating more accessible, engaging, and intelligent cultural spaces.

\section{Project Objectives and Relevance}
The project responds to the need to enhance access to small museums, which often lack adequate accessibility features due to their structure and location. Robots can assist or supplement museum staff, offering engaging experiences particularly appealing to younger audiences. In addition, they can provide support for foreign visitors, as well as for the visually and hearing impaired, through multilingual spoken explanations, text alternatives, sign language, and multimedia content. Moreover, telepresence robots can allow remote visits for individuals in nursing homes or isolated communities, facilitating reconnection with their places of origin.

This need was identified through reports such as Istat's ``Accessibility of Museums and Libraries" (2021)\footnote{https://www.istat.it/comunicato-stampa/laccessibilita-di-musei-e-biblioteche-anno-2021/}, highlighting widespread barriers in Italian museums and a lack of services for sensory and cognitive disabilities, particularly in less urbanized regions. Since the 1990s, robotics has been trialed in museums to support visitor orientation and engagement \cite{Thrun2000}, while telepresence robots have proven effective for social integration and interaction across distances \cite{Yousif2021, Kristoffersson2013}. The research challenges include emotional personalization, multi-user interaction, and the enhancement of shared experiences between remote and local participants \cite{Casoni2020}.

The project will be articulated in the following phases:
\begin{itemize}
    \item \textbf{Phase 1 – Requirement Analysis:} Benchmarking museum accessibility solutions and robotics in cultural spaces. User studies (questionnaires, interviews, focus groups) to assess preferences in the two main use cases. Contextual and technical constraints for the selected case studies.
    \item \textbf{Phase 2 – Design and Development:} Creation of an application ecosystem for telepresence booking and execution, and for onsite visitor interaction. Implementation of robot movement, dialogue, and multimedia storytelling capabilities.
    \item \textbf{Phase 3 – Human-Centered Design:} Co-design with stakeholders to integrate user perspectives into the project design. Ongoing usability and accessibility testing, both before and after the system release.
    \item \textbf{Phase 4 – Deployment and Monitoring:} Field deployment in the participating museums. Scheduled availability of onsite and remote services. Monitoring through quantitative and qualitative methods, including user interaction logs and satisfaction surveys.
\end{itemize}

\section{Impact}
The initiative will contribute to the economic, cultural, and social development of the territories involved, countering overtourism by promoting lesser-known destinations and enabling year round visits, particularly important for mountain museums. The project framework will be transferable to other museums. It embodies inclusive design principles, supporting accessibility for users with mobility, sensory, or cognitive limitations, and fostering intergenerational and intercultural participation.


\section{Evaluation and Expected Results}
The project will apply formative and summative evaluations with stakeholders, using metrics such as visit frequency, telepresence usage, interaction logs, and user feedback. Validated instruments like the Perceived Empathy of Technology Scale will assess emotional and usability dimensions \cite{schmidmaier2024perceived}. Qualitative feedback will be gathered through interviews with museum managers. Social media engagement will be tracked with analytics tools.

The expected beneficiaries include:
\begin{itemize}
    \item 17,000 general visitors,
    \item 2,400 people with disabilities,
    \item 6,800 foreign visitors,
    \item 2,000 students.
\end{itemize}

The deployment of ROBSO-PM will enhance accessibility, visitor engagement, and cultural connection—particularly among underserved audiences. The robot's presence is expected to attract new visitors and support inclusive cultural participation.

\section*{Declaration of Generative AI Use}
The authors declare that generative artificial intelligence (AI) was employed during the preparation of this manuscript solely for the purpose of language translation.

\bibliographystyle{IEEEtran}
\bibliography{biblio}

\end{document}